\documentclass[11pt]{article}

\usepackage[]{acl}  
\usepackage{times}
\usepackage{latexsym}

\usepackage[T1]{fontenc}
\usepackage[utf8]{inputenc}

\usepackage{microtype}
\usepackage{inconsolata}

\usepackage{graphicx}
\usepackage{booktabs}
\usepackage{array}
\usepackage{lscape}
\usepackage{subfigure}

\usepackage{amsmath}

\usepackage{amsmath,amsfonts,bm}

\def\eqref#1{equation~\ref{#1}}

\def\1{\bm{1}}

\DeclareMathAlphabet{\mathsfit}{\encodingdefault}{\sfdefault}{m}{sl}
\SetMathAlphabet{\mathsfit}{bold}{\encodingdefault}{\sfdefault}{bx}{n}

\usepackage{textcomp}
\usepackage{hyperref}
\usepackage{url}

\usepackage{longtable}
\usepackage{booktabs}

\title{Scaling Laws for Code: A More Data-Hungry Regime}

\author{
    \textbf{Xianzhen Luo}\textsuperscript{1,†}, 
    \textbf{Wenzhen Zheng}\textsuperscript{2,†}, 
    \textbf{Qingfu Zhu}\textsuperscript{1,*}, 
    \textbf{Rongyi Zhang}\textsuperscript{1}, \\
    \textbf{Houyi Li}\textsuperscript{\textbf{3}}, 
    \textbf{Siming Huang}\textsuperscript{\textbf{3}}, 
    \textbf{Yuantao Fan}\textsuperscript{\textbf{4}}, 
    \textbf{Wanxiang Che}\textsuperscript{\textbf{1,*}}\\
    \textsuperscript{1} Harbin Institute of Technology \ \
    \textsuperscript{2} Chinese Academy of Sciences \\
    \textsuperscript{3} Fudan University \ \
    \textsuperscript{4} Beijing University of Posts and Telecommunications \\
    \texttt{\{xzluo, qfzhu, car\}@ir.hit.edu.cn} \\
}

\begin{document}
\maketitle
\renewcommand{\thefootnote}{\fnsymbol{footnote}}
\footnotetext[1]{Corresponding author.}
\footnotetext[2]{Equal contribution.}

\begin{abstract}
Code Large Language Models (LLMs) are revolutionizing software engineering. However, scaling laws that guide the efficient training are predominantly analyzed on Natural Language (NL). Given the fundamental differences like strict syntax between code and NL, it is unclear whether these laws are directly applicable to code. To address this gap, we conduct the first large-scale empirical study of scaling laws for code, comprising 117 experimental runs with model sizes from 0.2B to 3.8B and training tokens from 2B to 128B. We fit the Chinchilla law and the Farseer law. First, the results show that the more expressive Farseer law offers greater accuracy. Second, the analysis reveals that Code LLMs scale effectively with model size. Crucially, code represents a more data-hungry regime, requiring a substantially higher data-to-parameter ratio than NL. Finally, two additional sets of experiments on code-NL mixtures show that NL benefits resource-constrained scenarios, but becomes a detriment at higher compute budgets.

\end{abstract}

\section{Introduction}

Code Large Language Models (LLMs) trained on large-scale code corpora have achieved remarkable code-related capabilities \citep{Chen2021EvaluatingLL,Li2022CompetitionLevelCG,Roziere2023CodeLO,Li2023StarCoderMT,huang-etal-2025-opencoder,openaiCompetitiveProgrammingLarge2025,huiQwen25CoderTechnicalReport2024}.
Built upon these LLMs, applications such as OS, GUI, and Terminal Agents significantly enhance developer productivity and substantially impact the field \citep{hu2025agents,tang2025surveymllmbasedguiagents,tbench_2025}. 
The fuel for this technological revolution is the continuous growth of data and model size, which also incurs substantial computational costs \citep{Kaplan2020ScalingLF,Hoffmann2022TrainingCL}. 
Training frontier LLMs requires thousands of petaflop/s-days and millions of dollars, making it impractical to conduct ablation experiments on the largest models, whether on the structure, data, or training strategies~\cite{brown2020language,deepseek-aiDeepSeekV3TechnicalReport2025,yangQwen3TechnicalReport2025a}.

\begin{figure}[t!]
\centering
\includegraphics[width=\columnwidth]{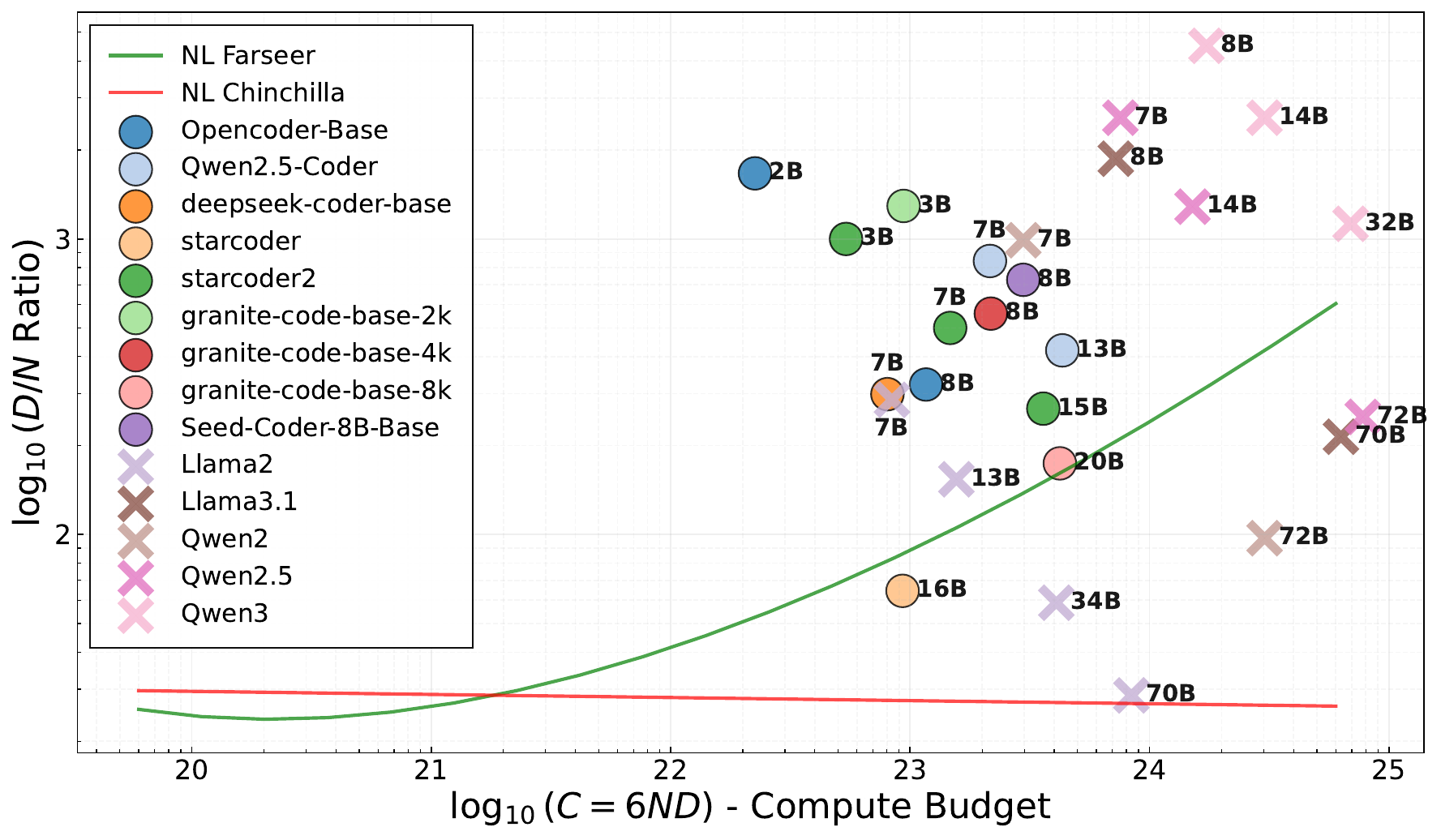}
\caption{The relationship between compute budget ($C$) and the optimal data-to-parameter ratio ($D/N$) for various LLMs. The Farseer and Chinchilla curves are derived from natural language data. Code LLMs is represented as circles while general LLMs is cross.}
\label{fig:intro_plot_1}
\end{figure}
Scaling laws provide a theoretical foundation to address this challenge \citep{Kaplan2020ScalingLF,Hoffmann2022TrainingCL}. By using results from smaller models to fit empirical formulas, scaling laws describe the relationship between model performance (typically the validation loss, $L$) and factors like model size ($N$), dataset size ($D$), and compute ($C$). 
They provide important references of resource allocation on the data and model for training general-purpose natural language (NL) LLMs~\cite{brown2020language,team2025kimi}.
However, code, as a highly structured data type, has statistical properties that fundamentally differ from NL: it has strict syntax, complex long-range dependencies, and unique vocabulary distributions~\cite{allamanis2018survey}. 
This raises the question: \textit{Can scaling laws developed for NL be applied to code?}

Exploring this open question is particularly important. 
As shown in Figure~\ref{fig:intro_plot_1}, Code LLMs tend to be concentrated in smaller sizes, with most prominent models being 32B or smaller. 
\textit{Does model scaling for Code LLMs plateau more quickly?} 
Futhermore, Code LLMs often require a larger data-to-parameter ($D/N$) ratio, which Figure~\ref{fig:intro_plot_1} shows deviates significantly from the predictions of NL scaling laws. 
\textit{Does this imply that Code LLMs fundamentally require more code data?} 
Answering these questions is crucial for understanding and guiding the future development of code pretraining.

To address these questions, we conduct the first systematic explorations of scaling laws specifically for code. Our study comprises 117 experiments with $N$ ranging from 0.2B to 3.8B and $D$ from 2B to 128B tokens. 
These models are trained on a curated public code corpus, using a consistent architecture and optimized hyperparameters for each run, and their losses are evaluated on a high-quality, held-out validation set.
By fitting these results to both the Chinchilla and Farseer formulations, we validate that the more expressive \textbf{Farseer law accurately models code's scaling behavior}, achieving a significantly better fit on experiment runs and higher predictive precision on validation points.
The fitted Farseer surface reveals a key dynamic: \textbf{contrary to initial expectations, Code LLMs scale robustly with $N$}.
Further analysis of the compute-optimal $D/N$ ratio reveals that \textbf{code is significantly more ``data-hungry'' than NL}. For a given $N$, a Code LLM requires substantially more data to reach its optimal performance. We attribute this to the inherent repetitiveness of code. Massive token volumes are needed to introduce new applications.
This insight provides a compelling explanation for the smaller size of typical code LLMs: it is likely a consequence of the scarcity of high-quality code data and practical low-latency inference requirements like code completion, rather than a lack of benefit from scaling model parameters.

Given the relative abundance of NL data and some practice of training Code LLMs on code-NL mixtures~\cite{lozhkov2024starcoder2,Li2023StarCoderMT}, we also investigate the potential to augment code model training with NL. 
We conduct a total of 234 additional experiments, comprising a full 117-run sweep for each of the two mixture ratios (70\%/30\% and 30\%/70\% code-NL).
The results indicate that when the volume of code data is limited, incorporating a moderate amount of NL data can indeed enhance performance on code. However, this benefit diminishes and eventually becomes a detriment as the proportion of NL data increases or as more pure code data becomes available.

Our contributions are summarized as follows:
\begin{itemize}
\item To our knowledge, we conduct the first scaling law for code, demonstrating that its loss is highly predictable and can be accurately modeled by an existing law formulation.
\item We find that Code LLMs exhibit good scaling properties with respect to $N$. The compute-optimal D/N ratio for code is significantly larger than for NL.
\item We reveal a reversal of effects for code-NL mixtures: NL provides a performance boost in low-compute settings but degrades performance at larger scales, providing clear guidance on when to use mixed-data strategies.

\end{itemize}

\section{Methodology}

\subsection{Background and Motivation}
Scaling laws are empirical formulas that guide the efficient training of LLMs by describing the relationship between $L$ and key factors like $N$, $D$, and $C$. 
Foundational work focused primarily on NL, such as the Chinchilla law~\citep{Hoffmann2022TrainingCL}, proposed a simple power-law relationship: 
\begin{equation}
L(N, D) = E + \frac{A}{N^a} + \frac{B}{D^b}
\end{equation}.
While powerful, its constant exponents $(a,b)$ mean $N$ and $D$ are scaled proportionally, failing to capture the nuanced interplay where larger models learn more efficiently from data.
To address this, a recent refinement, the Farseer law~\citep{Li2025Farseer} has introduced more expressive formulations, where the scaling exponents are themselves functions of $N$, such as:
\begin{equation}
L(N,D)=e^{s \cdot N^q+S}+e^{B \cdot N^b + Q} \cdot D^{-e^{A \cdot N^a + E}}
\end{equation}.
Possessing such a law enables the calculation of the compute-optimal $D/N$ ratio for any fixed $C$, thereby guiding efficient resource allocation.

However, while these laws have proven effective for NL, their direct applicability to code is questionable. Code possesses fundamentally different statistical properties, including strict syntax and complex long-range dependencies~\cite{allamanis2018survey}. 
Beyond these theoretical distinctions, empirical observations suggest that NL-derived laws are a poor fit.
As illustrated in Figure~\ref{fig:intro_plot_1}, the D/N ratios of prominent code LLMs deviate significantly from the optimal values predicted by NL scaling laws.
This discrepancy highlights the need for a dedicated investigation to establish a scaling law tailored specifically for code, providing more accurate guidance for training future Code LLMs.

\subsection{Experimental Design}
Our experimental design is a budget-conscious adaptation of the comprehensive methodology from Farseer~\citep{Li2025Farseer}. Given that a full replication is computationally prohibitive, our design focuses on a strategic selection of configurations that are optimized to balance deep informational gain with a feasible computational cost.

\begin{figure}[t]
\centering
\includegraphics[width=\columnwidth]{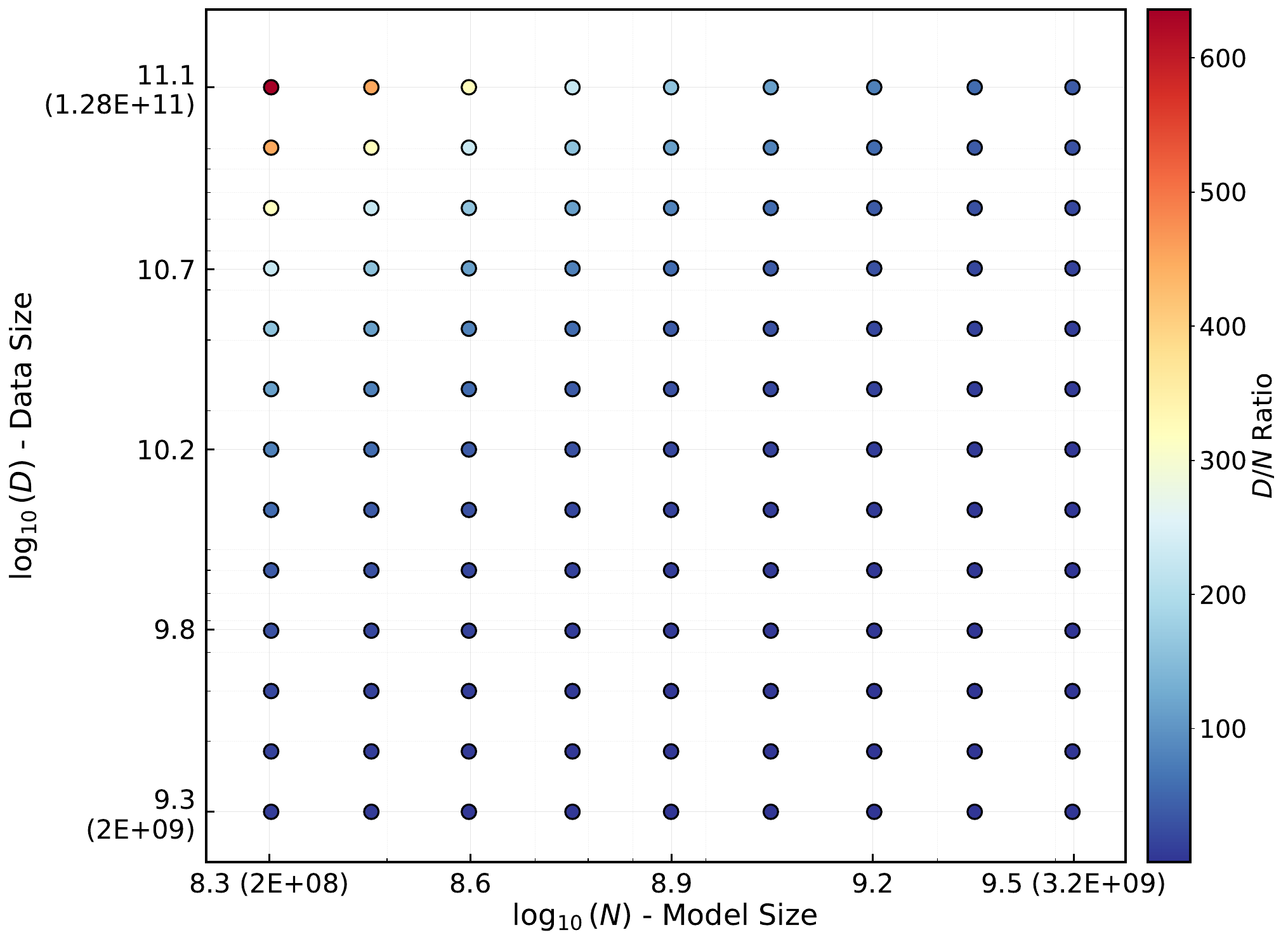}
\caption{
A scatter plot of our 117 experimental points (colored by $D/N$ ratio), showing model parameters ($N$) versus training tokens ($D$). 
}
\label{fig:dn_scatter}
\end{figure}

\textbf{Sampling Strategy.}
Given the significant computational cost of training, we carefully curate a set of 117 configurations by sampling log-uniformly across $N$ and $D$. 
We constrain $N$ to a range of 0.2B to 3.8B. The lower bound is chosen because models smaller than 0.2B contribute minimally to fitting, while the upper bound considers the current size distribution of Code LLMs. Furthermore, we prune extreme $D/N$ ratios, including both very high and very low, from the original Farseer design. This retains $D$ in a range of 2B to 128B. Figure~\ref{fig:dn_scatter} visualizes the distribution across $N$, $D$, and $D/N$ ratio of the final 117 experimental points.

\textbf{Model Architecture and Parameterization.}
To strictly isolate the effects of scale, all 117 models in our study share a consistent decoder-only Transformer architecture. This architecture incorporates modern, effective components for large-scale training, including SwiGLU activations, Rotary Position Embeddings, and RMSNorm~\citep{shazeer2020glu,su2021roformer,zhang2019root}. To determine the specific hyperparameters such as model dimension, number of attention heads, and layer count for each model, we employ the deterministic procedure from Farseer~\citep{Li2025Farseer}. This ensures that all models are well-proportioned by maintaining near-optimal architectural aspect ratios, making them both structurally efficient and directly comparable. A complete table of all model configurations is available in Appendix~\ref{app:model_configs}.

\textbf{Training Hyperparameters.}
To ensure that each model is trained efficiently without a cost-prohibitive hyperparameter search, we follow StepLaw's optimal hyperparameter scaling rules to set near-optimal learning rates, global batch sizes (GBZ) from $N$ and $D$~\citep{Li2025PredictableSP}. 
Crucially, StepLaw validates its scaling rules specifically on a code pretraining recipe, confirming the applicability to our study.
We use the AdamW optimizer~\citep{loshchilov2019decoupled} with a cosine learning rate decay schedule for all experiments.

\textbf{Train Set and Validation Set.}
We utilize the high-quality code pre-training data from Opencoder~\cite{huang-etal-2025-opencoder}, which is sourced from public GitHub repositories and has undergone meticulous deduplication and filtering. To create a more balanced programming language distribution, we downsample the initial 1216.93B token corpus, resulting in our final 895.51B token train set.
For evaluation, our validation set is a 6.3M token internal codebase from~\citet{xuyang2025compressionreallylinearcode}. 
This dataset consists of production-grade code used in real-world applications, ensuring its practical relevance, and has been rigorously checked to guarantee no overlap with the train set.
Full details about the two set are provided in Appendix~\ref{app:data_details}.

\textbf{Computational Setup.} 
All experiments are conducted on a cluster of NVIDIA H100-80GB GPUs. Once the GBZ is determined, we configure the number of GPUs and the per-GPU micro batch size by managing the trade-off between minimizing runtime (which favors more GPUs) and maximizing Model FLOPs Utilization (MFU, which favors fewer, more saturated GPUs). 
The final configurations for each run utilized between 8 and 128 GPUs. 
The total compute expended is approximately 13,600 H100 GPU-days. Further scheduling details are in Appendix~\ref{app:comp_setup}.

\section{Experiment Results}

\begin{figure*}[h]
\centering
\includegraphics[width=\textwidth]{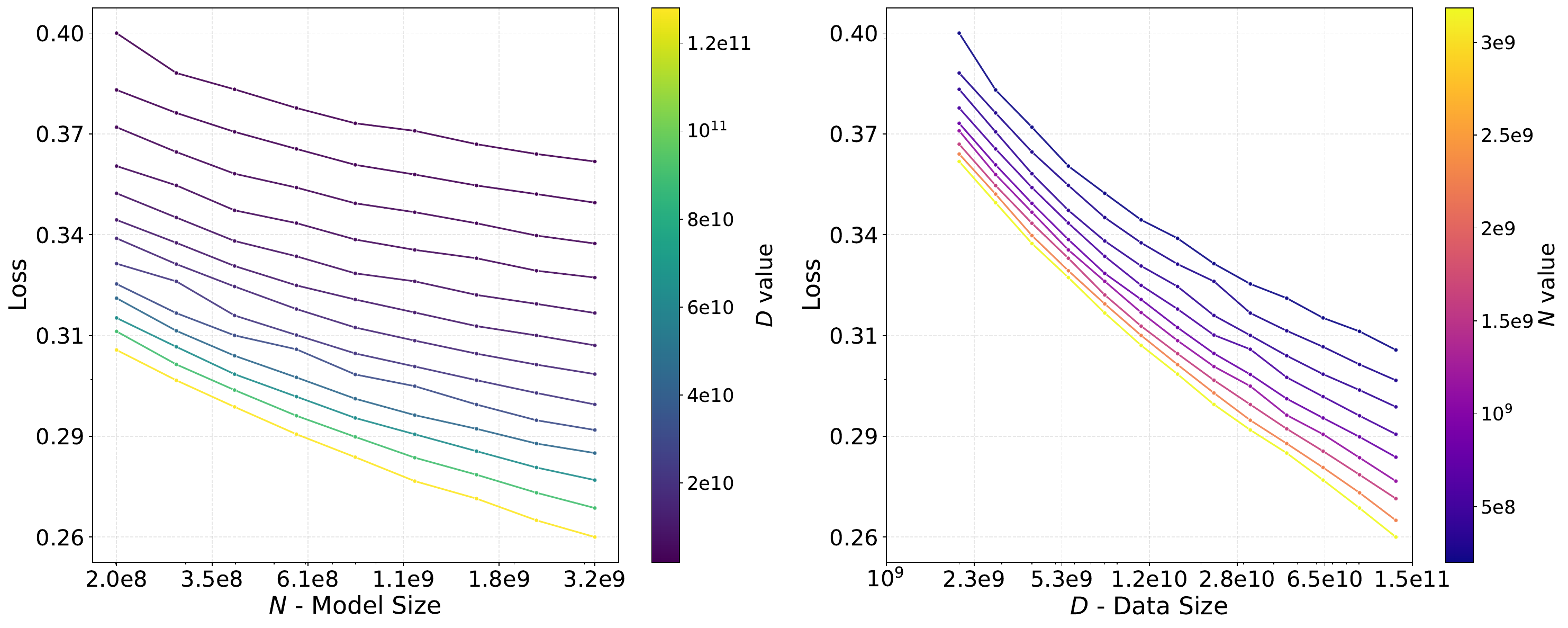}
\caption{Scaling behavior of loss across different model sizes ($N$) and data sizes ($D$). (a) Loss decreases with increasing $N$ for fixed $D$ values. (b) Loss decreases with increasing $D$ for fixed $N$ values. The color gradients represent different $D$ values (left) and $N$ values (right). Both plots use double-logarithmic axes.}
\label{fig:bpc_heatmap}
\end{figure*}

\begin{table*}[t]
\centering
\setlength{\tabcolsep}{3pt}
\renewcommand{\arraystretch}{0.95}
\begin{tabular}{ccccccccccc}
\toprule
\textbf{D/N} & \textbf{N (B)} & \textbf{D (B)} & \textbf{GBZ} & \textbf{GPUs} & \textbf{MBZ} & \textbf{PL\textsubscript{F}} & \textbf{PL\textsubscript{C}} & \textbf{Loss}  & \textbf{RE\textsubscript{F}(\textperthousand)} & \textbf{RE\textsubscript{C}(\textperthousand)}\\
\midrule
20 & 6.37 & 127 &  640 & 160 & 2 & 0.259271 & 0.265707 & 0.256833 & \textbf{9.49} & 34.55\\
150 & 2.27 & 341 &  1080 & 120 & 9 & 0.253488 & 0.262330 & 0.253786 & \textbf{1.17} & 33.67 \\
424 & 1.34 & 567 &  1456 & 112 & 13 & 0.255846 & 0.262939 & 0.258546 & \textbf{10.44} & 16.99 \\
\bottomrule
\end{tabular}
\caption{
Validation results for three model configurations trained at a large compute budget.
\textbf{GPUs} indicates the number of GPUs used for the run. \textbf{GBZ} and \textbf{MBZ} refer to the Global and Micro Batch Size. 
\textbf{PL\textsubscript{F}} and \textbf{RE\textsubscript{F}} denote the predicted loss and relative error from Farseer, respectively, with \textbf{PL\textsubscript{C}} and \textbf{RE\textsubscript{C}} representing the counterparts for Chinchilla.
\textbf{Loss} is the empirically measured validation loss. 
}
\label{tab:val_point}
\end{table*}

\subsection{Scaling Behavior}

After training, the validation loss is computed for each of the 117 models. As a prerequisite for fitting a scaling law, we first qualitatively verify that the model performance scales predictably with $N$ and $D$. Figure~\ref{fig:bpc_heatmap} provides a comprehensive visualization of this behavior. Figure~\ref{fig:bpc_heatmap}(a) shows that for a fixed $D$, the validation loss monotonically decreases as $N$ increases, a trend that holds consistently across all tested $D$. Symmetrically, Figure~\ref{fig:bpc_heatmap}(b) demonstrates that for any given $N$, the loss also smoothly declines as $D$ grows. The near-linear trends on these log-log plots indicate approximate power-law relationships between loss and $N$/$D$, consistent with scaling-law assumptions. These clear and consistent trends suggest that performance on code exhibits stable scaling behavior, thus providing an empirical foundation for our subsequent quantitative analysis.

\subsection{Scaling Law Fitting}
To quantitatively model the observed scaling behavior, we fit our 117 experimental data points to the Chinchilla and the Farseer, yielding the specific formulations shown in Equation~\ref{eq:code_chinchilla} and Equation~\ref{eq:code_farseer}:

\begin{equation}
L(N, D) = 0.2193 + \frac{534.374}{N^{0.4853}} + \frac{76.0743}{D^{0.2983}}
\label{eq:code_chinchilla}
\end{equation}

\begin{equation}
\begin{split}
L(N,&D) = \exp(-0.0047 \cdot N^{0.239} - 0.8188) \\
&+ \exp(62.8936 \cdot N^{-0.0614} - 14.0414) \\
&\cdot D^{-\exp(-0.0209 \cdot N^{0.1943} - 0.1826)}
\end{split}
\label{eq:code_farseer}
\end{equation}


The Farseer formulation provides a better fit to our data, exhibiting a lower mean relative error than the Chinchilla model (0.82‰ vs. 1.03‰). 
This more accurate law also makes a fundamentally different asymptotic prediction: an irreducible loss of zero, in contrast to the 0.2 limit from Chinchilla, suggesting there may be no theoretical performance ceiling for scaling (Details in Appendix~\ref{app:irreducible_loss}). However, the ultimate test of a scaling law includes its power to extrapolate and predict performance in new, large-scale scenarios, which motivated our dedicated validation experiment.


\subsection{Scaling Law Validation}
\label{sec:scaling_law_val}
Within available computational resources, we design a rigorous validation experiment at a fixed large compute budget of $C=5.36\times10^{21}\text{FLOPs}$. 
Three distinct D/N ratios are selected:
(1) D/N=20, the established optimum for NL Chinchilla, to test the direct applicability of NL-derived heuristics; (2) D/N=150, the predicted optimum for code from Farseer (Equation~\ref{eq:code_farseer}), it is also really close to Chinchilla law's prediction thus run once for efficiency; and (3) D/N=424, an extreme data-heavy as a upper-bound point to evaluate the extrapolation capabilities of equations.

\begin{figure}[]
\centering
\includegraphics[width=\columnwidth]{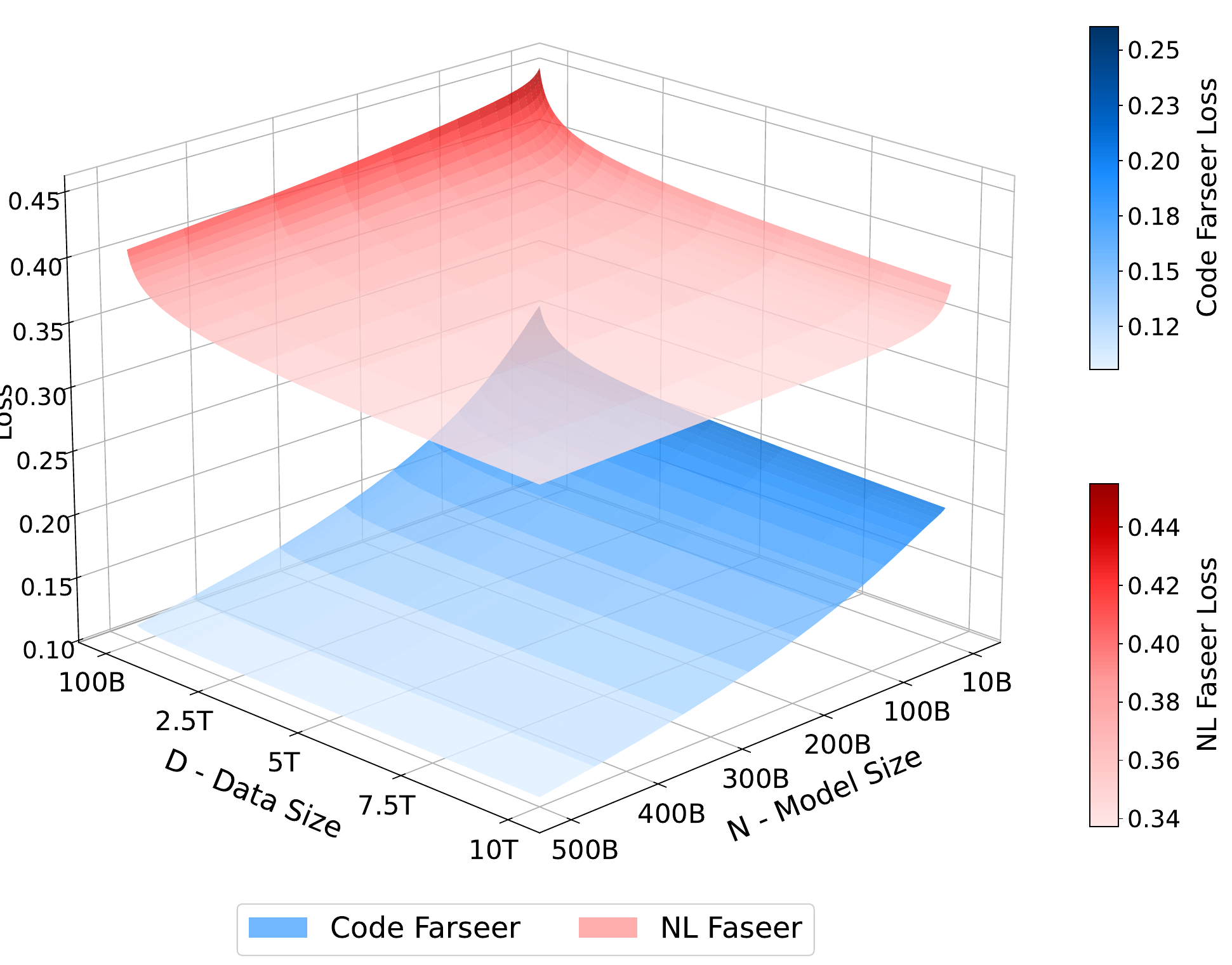}
\caption{3D visualization of the fitted Farseer scaling-law surfaces for Code (blue) and Natural Language (orange) over $(N,D)$.}
\label{fig:farseer_surface}
\end{figure}

Table~\ref{tab:val_point} presents the predictions from both equations alongside the empirical results and offers two primary findings.
First, optimal training derived from NL is suboptimal for code.
This is evidenced by the empirical losses, where the model trained at $D/N=150$ achieves a significantly lower loss than the one at the NL-optimal ratio of $D/N=20$. However, this does not imply an infinite appetite for data, as the loss at $D/N=424$ is higher again.
Second, Farseer demonstrates better predictive accuracy. It outperforms Chinchilla across all three validation points, achieving a remarkably low relative error of just 1.17\textperthousand at the empirically confirmed optimum. 
The prediction errors at the extreme D/N ratios, while still low at approximately 1\%, are relatively higher.
This is likely because such settings are less common in practice and thus are less represented in our initial 117 fitting points. 

In summary, our experiments demonstrate that while NL training recipes are suboptimal for code, code performance is highly predictable under the Farseer formulation. 
We emphasize that our primary goal is to identify the most suitable \textit{form} of the scaling law for code, rather than providing universal parameters, as the specific values are inherently tied to the experimental setup and may vary subtly across different implementations

\section{Analysis}
Having established in Section~\ref{sec:scaling_law_val} that the Farseer law provides an accurate model for code scaling, we now use this formulation to analyze the underlying dynamics and derive key insights.

\subsection{Scaling Surface}
Figure~\ref{fig:farseer_surface} shows the fitted Farseer surfaces for Code vs. NL over $(N,D)$. The visual inspection reveals some fundamental differences.

\textbf{Code Has Lower Intrinsic Loss.} The validation losses for code and NL models are computed on their own validation sets. In our fitted surfaces, code loss is generally lower than NL across a broad range of $(N, D)$. This suggests a difference in intrinsic entropy between the two modalities. Although code can be logically complex, its statistical regularities, like strict syntax, standard templates, keywords, and common programming idioms, are often more prevalent than in NL~\cite{10.1609/aaai.v38i1.27798}. This view aligns with evidence from speculative decoding, where code tokens are easier to predict than NL tokens, enabling higher acceleration~\citep{leviathan2023fast}. Differences across programming languages also appear smaller than across NL, making code tokens easier to learn and predict~\cite{Roziere2023CodeLO,hasan2025assessing}.

\textbf{Code Exhibits More Sustained Scalability.} Although code has a lower absolute loss, its surface has still sustained steepness. As visualized in Figure~\ref{fig:farseer_surface}, while the NL surface begins to visibly flatten at larger scales, the code surface maintains a more consistent and steep downward gradient across the entire range. This indicates that, within our experimental scope, code models are not approaching a saturation point. Instead, they continue to derive significant performance benefits from further increases in either $N$ or $D$, highlighting the vast potential for scaling even larger Code LLMs.


\textbf{Code Has Different Scaling Properties.} Contrary to our initial hypothesis that code would plateau faster along $N$, we observe excellent scaling along the $N$ axis: gains from increasing $N$ are larger for code than NL. Whereas for $D$, achieving a comparable reduction in loss requires a substantially larger volume of additional data. 
This asymmetry may be attributed to the fundamental structure of code. Code is governed by a finite set of syntactic rules and structural patterns~\cite{winskel1993formal}. On one hand, 
a bigger model can better understand how basic programming constructs combine to form complex algorithms~\cite{shi2024can,liu2024systematic}. 
On the other hand, because these underlying rules are repetitive across the corpus, a vast number of new tokens is needed to encounter genuinely novel application scenarios, leading to less efficient $D$-scaling~\cite{casalnuovo2019studying}.



\begin{figure}[t]
\centering
\includegraphics[width=\columnwidth]{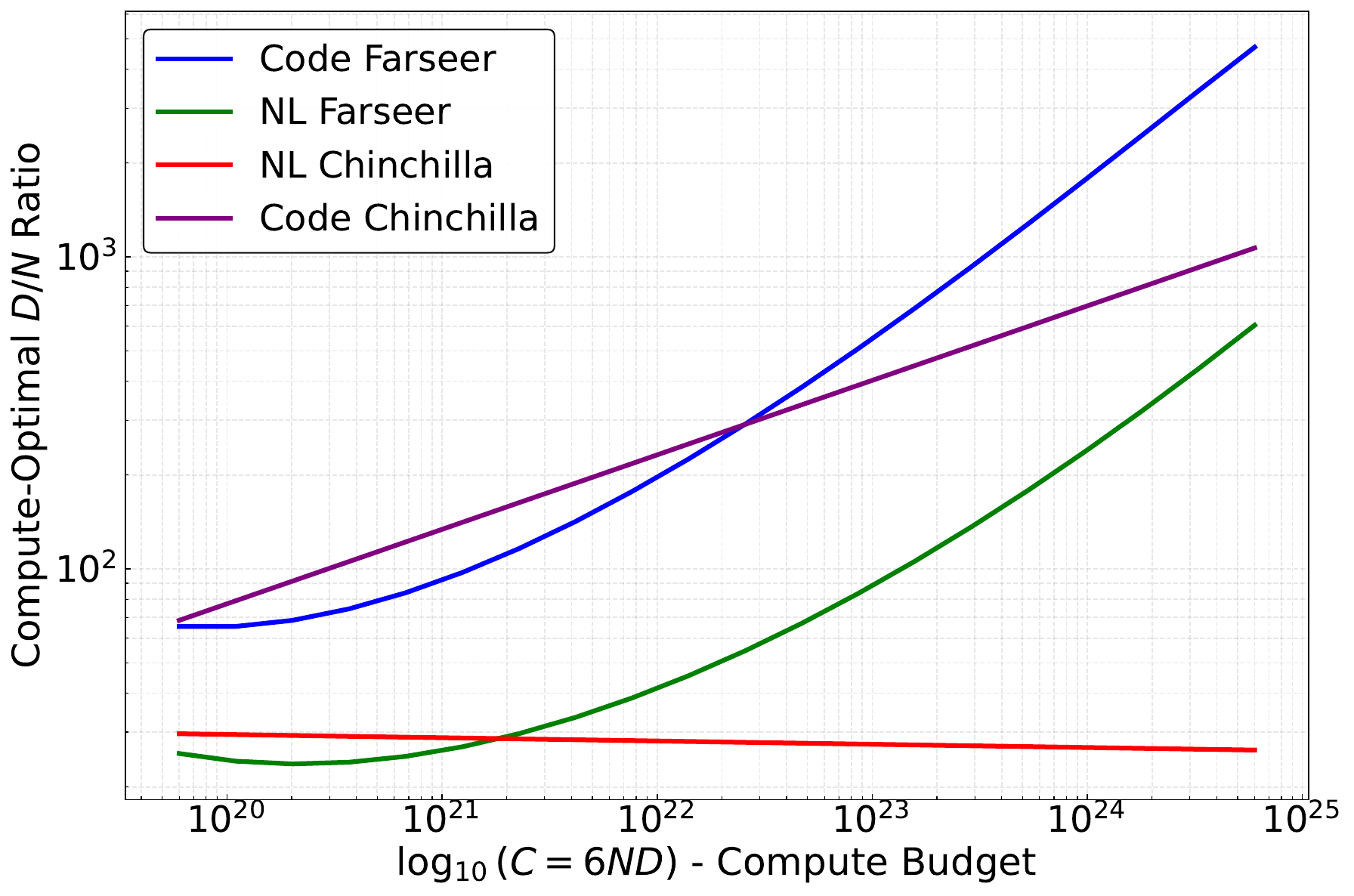}
\caption{The compute-optimal $D/N$ ratio vs. compute budget $C$. }
\label{fig:optimal_dn}
\end{figure}

\subsection{Optimal D/N Ratio}
\label{sec:optimal_dn}

Based on the scaling landscapes, we derive the optimal $D/N$ ratio as a function of $C$ in Figure~\ref{fig:optimal_dn}. 

\textbf{Code LLMs Tend to Be More Data-Hungry.} Across the compute budgets we studied, the optimal D/N ratio for code is consistently higher than for NL, and this gap widens with scale. We attribute this to the lower average information density of large-scale, conventional code corpora, such as GitHub data used in our training. 
Due to the inherent repetitiveness, the model masters common patterns in the early stages of training.
While further convergence can be achieved by stacking more data by expanding to cover more scenarios, the process is highly costly. 
Therefore, we posit that a more effective path to improvement lies in enhancing data quality, for example, through targeted upsampling of high-complexity code, rather than just blindly increasing data quantity.

\begin{figure}[]
\centering
\includegraphics[width=\columnwidth]{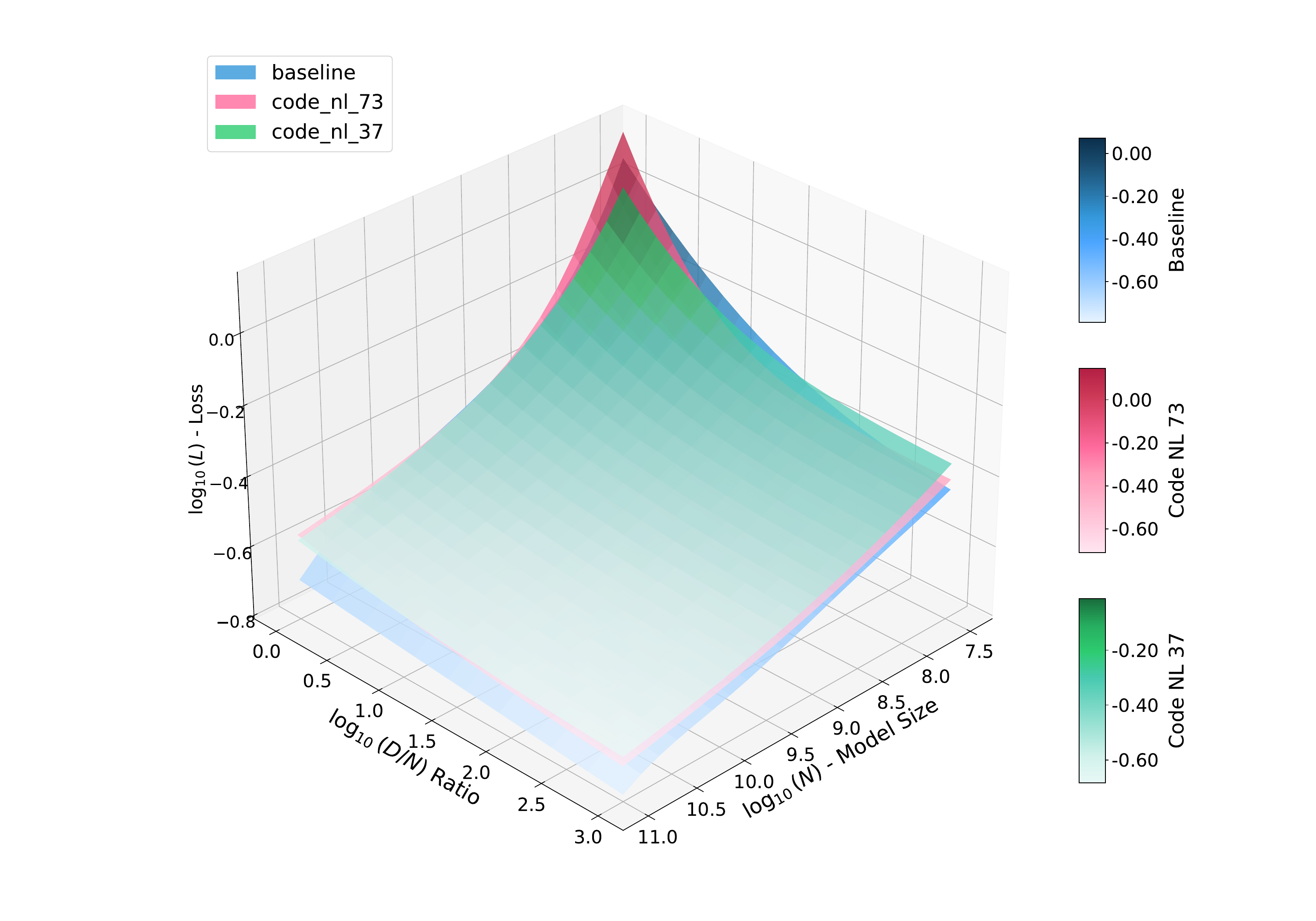}
\caption{3D visualization comparing scaling surfaces for different data mixtures: baseline (100\% code), \texttt{code\_nl\_73} (70\% code + 30\% NL), and \texttt{code\_nl\_37} (30\% code + 70\% NL). $D/N$ ratio only considers the number of code tokens.}
\label{fig:data_mixing_3d}
\end{figure}

\textbf{The Optimal $D/N$ Ratio Grows Exponentially with $C$.}
The optimal $D/N$ ratio for code is not static but grows progressively with $C$.
While NL Chinchilla proposes a near-constant $D/N$ ratio, 
Figure~\ref{fig:optimal_dn} reveals that even the Chinchilla law fit on code data exhibits a clear upward trend.
This growth is linear because Chinchilla assumes that $N$ does not affect the dynamics of $D$. 
However, Farseer better captures the scaling dynamics and posits that a larger $N$ accelerates learning from $D$, thus demanding more data for convergence. 
This results in a super-linear growth of the optimal $D/N$ ratio, confirming that the demand for data accelerates at higher compute scales.



\textbf{D/N May Help Explain Smaller SOTA Code LLMs.} 
This high data requirement helps explain why mainstream code LLMs are generally smaller than their NL counterparts. The primary bottleneck appears to be the lack of sufficient data to optimally train a massive model, rather than diminishing returns from scaling the model size itself. In practice, model size is further constrained by the limited availability of code data and by factors like inference cost, especially for low-latency applications like code completion. Nevertheless, larger Code LLMs can still deliver significant performance benefits when sufficient data and compute are available like Qwen3-Coder~\cite{yangQwen3TechnicalReport2025a}.

\begin{figure*}[]
\centering
\includegraphics[height=0.4\textheight]{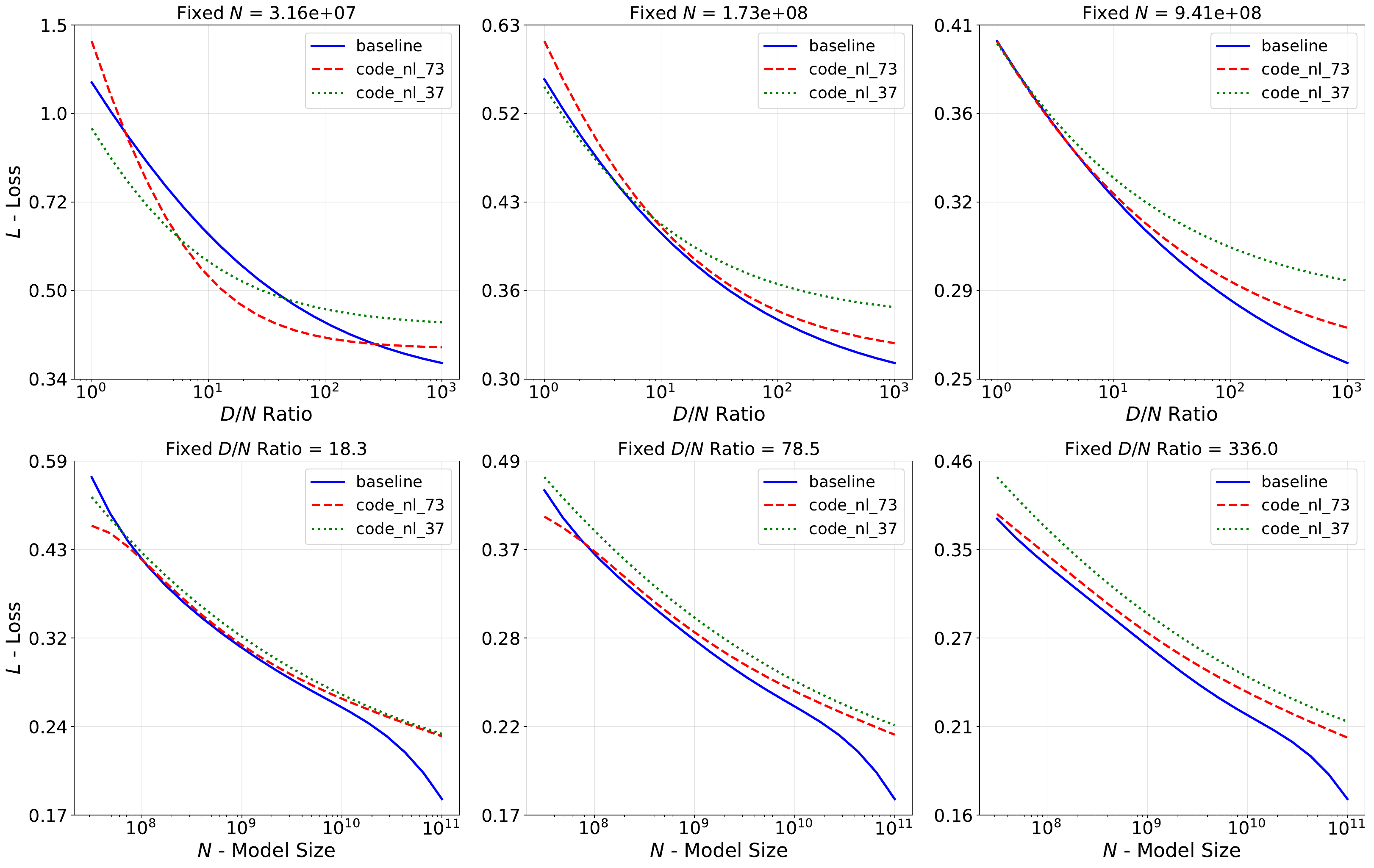}
\caption{2D slices of three scaling law surfaces. Top Row: Loss as a function of the $D/N$ ratio for three fixed $N$. Bottom Row: Loss as a function of $N$ for three fixed $D/N$ ratios. $D/N$ ratio only considers the number of code tokens and does not include the NL data.}
\label{fig:data_mixing_fixed_n}
\end{figure*}

\subsection{Data Mixing}

Section~\ref{sec:optimal_dn} reveals that Code LLMs are significantly more data-hungry, a practical challenge given the relative scarcity of high-quality code data. 
Furthermore, while it is common industry practice to train on proprietary code-NL mixtures~\cite{Li2023StarCoderMT,lozhkov2024starcoder2}, the impact of these ratios on scaling behavior is largely unexplored.
These considerations motivate our investigation into whether NL corpora can be strategically leveraged to improve code model performance.

To this end, our computational budget allows for two additional full experimental sweeps, for which we select two distant points on the mixture spectrum to maximize the observable differences: (1) \textbf{70\% Code + 30\% NL}: A code-dominant mixture, denoted as \texttt{code\_nl\_73}. (2) \textbf{30\% Code + 70\% NL}: An NL-dominant mixture, denoted as \texttt{code\_nl\_37}.
For each mixture, we replicate our full set of 117 experimental runs, covering the same range of $N$ and $D$ as the baseline. This rigorous approach allows us to fit a new, distinct Farseer law for each condition, enabling a direct, comprehensive and fair comparison of their scaling behaviors. 
To ensure this comparison is practical, it is crucial to note that Figure~\ref{fig:data_mixing_3d} and Figure~\ref{fig:data_mixing_fixed_n} define $D$ and the $D/N$ ratio using only the number of code tokens, not the total token count of mixtures.


Figure~\ref{fig:data_mixing_3d} compares the fitted Farseer surfaces for baseline and two data mixtures, revealing a scale-dependent trade-off. 
In the low-compute regime (i.e., at smaller $N$ and $D/N$ ratios), 
the surfaces have a crossover phenomenon. Particularly, the \texttt{code\_nl\_73} can achieve a lower loss than the pure code baseline.
However, as the scale of computing increases, the baseline quickly becomes and remains superior.
Furthermore, the baseline surface maintains a steeper and more consistent downward gradient. In contrast, the surfaces for the mixtures begin to show signs of saturation, with their rate of loss reduction diminishing more quickly.

To further dissect the scaling dynamics of the data mixtures, we plot 2D slices of the 3D surfaces in Figure~\ref{fig:data_mixing_fixed_n}. These slices allow for a more granular analysis of performance under controlled conditions, yielding several key insights. 
The top panels show that the benefit of NL data is most pronounced for smaller models. 
The top-left panel (Fixed $N=31.6$M) provides clear evidence of the crossover phenomenon suggested by the 3D surfaces. At low $D/N$ ratios (e.g., < 40), both the \texttt{code\_nl\_73} and \texttt{code\_nl\_37} achieve lower loss than the baseline. 
This empirically validates that when code data is limited, a moderate amount of NL data can indeed enhance a smaller model's performance on code tasks.
We hypothesize this is because the limited in-domain data is insufficient to learn robust representations, and the diverse NL data acts as a powerful regularizer and provides transferable world knowledge~\cite{Chen2021EvaluatingLL}. 
The \texttt{code\_nl\_73} is particularly effective, maintaining its advantage until the $D/N$ ratio approaches 300. The \texttt{code\_nl\_37}, while optimal in a very small and impractical $D/N < 5$ regime, quickly falls behind.
However, as $N$ increases to 173M and further to 0.94B (middle and right panels), the model's capacity to learn from pure code improves, and the mixtures' advantages almost completely vanish. 
At $N=0.94$B, the baseline is superior across all $D/N$ ratios.
For these larger models, which are high-capacity, the potential benefit of knowledge transfer from NL may be outweighed by the cost of distributional shift. The introduction of out-of-domain data can act as a performance impediment from the outset.



This pattern is even more pronounced when observing scaling with $N$ at fixed $D/N$ ratios in the bottom row plots of Figure~\ref{fig:data_mixing_fixed_n}. 
At a low, data-scarce ratio of $D/N=18.3$ (bottom-left panel), the \texttt{code\_nl\_73} mixture is again significantly stronger than the baseline for smaller models ($N < 0.1$B), reinforcing the hypothesis that NL data provides a crucial boost when in-domain data is insufficient. 
As the $D/N$ ratio increases to 78.5 and then to 336, the in-domain data becomes more and more. The need for the NL supplement diminishes, and its role shifts from a helpful regularizer to an out-of-domain distraction. 
Consequently, at $D/N=336$ (bottom-right panel), the three curves become stably stratified, with the baseline performing the best, followed by \texttt{code\_nl\_73}, and then \texttt{code\_nl\_37}, with no crossovers observed.




In summary, these 2D projections provide a clear prescription for the role of NL data in code pretraining. It acts as a valuable supplement in low-resource settings, where it likely provides a regularizing effect and transferable knowledge to smaller models trained on limited in-domain data. However, this benefit is a trade-off. As model capacity or the availability of in-domain data increases, the performance cost of the distributional shift outweighs the benefits, and pure, in-domain data becomes the unambiguous key to achieving optimal performance at scale.

\section{Related Work}

\textbf{Neural Scaling Laws.}
\citet{Kaplan2020ScalingLF} first established the power-law relationship between $L$ and factors like $C$, $N$, and $D$, sparking the industry's focus on aggressively scaling up model size.
This trend is refined by Chinchilla~\citep{Hoffmann2022TrainingCL}, which argued that many LLMs were significantly undertrained. They introduced the concept of compute-optimal scaling, proposing that $N$ and $D$ should be scaled proportionally. Subsequent research has expanded beyond these core variables to incorporate more fine-grained factors, including inference costs~\citep{touvron2023llama,su2024unraveling,10.5555/3692070.3693840}, Mixture-of-Experts (MoE) architectures~\citep{zhao2025towards,ludziejewski2025joint,tian2025towards}, data quality and distribution~\citep{shuai2024scaling,chen2025revisiting,qin2025scaling,liu2025regmix}, and the specific downstream capabilities~\citep{lin2024scaling,roberts-etal-2025-compute}. 
Most recently, Farseer~\citep{Li2025Farseer} challenged the fixed optimal $D/N$ ratio, demonstrating that it should increase with $C$ and proposing a more expressive formulation to capture this dynamic.

\textbf{Large Language Models for Code.}
Code LLMs have progressed rapidly alongside general-purpose LLMs. They are predominantly trained on corpora from GitHub, StackExchange, and other code-centric web sources. Early work continue-pretrain the NL LLMs to acquire code capabilities~\cite{Chen2021EvaluatingLL,chowdhery2023palm,anil2023palm,zhu2024deepseek,team2024codegemma,huiQwen25CoderTechnicalReport2024}. Subsequently, stronger Code LLMs are trained from scratch on pure code or code-heavy mixtures ~\cite{Li2022CompetitionLevelCG,Roziere2023CodeLO,nijkamp2023codegen,fried2023incoder,allal2023santacoder,guo2024deepseek,mishra2024granite,seed2025seed,huang-etal-2025-opencoder}.

Despite these advances, scaling laws for code remain poorly understood. Current code models are often trained on heuristic settings without a strong empirical basis. We first addresses this gap. We analyze models trained on pure code and controlled code-NL mixtures to establish a principled guide for the compute-optimal training of foundation models for code.

\section{Conclusion}


In this work, we conducted the first large-scale empirical study of scaling laws for code. We first find that the Farseer paradigm can accurately predict the scaling properties of code, which demonstrates that code adheres to existing scaling law frameworks. The analysis of scaling properties reveals that code scales robustly with model size, but exhibits a significant demand for data. This is further validated by the fact that the compute-optimal $D/N$ ratio for code is substantially higher than for NL, and that this gap widens as the compute budget increases. Finally, code-NL mixtures experiments reveal that while NL data can be beneficial in low-resource scenarios, pure code data remains superior as model size increases or when in-domain data is abundant.


\section*{Limitations}
While this study provides foundational insights into the scaling laws of code, it is subject to several limitations. First, due to computational constraints, our analysis is based on a curated set of 117 experimental points rather than the 400+ configurations used in the original Farseer study. Although our strategic selection covers the most critical regions, a larger set of fitting points could further refine the precision of our fitted law. Second, our validation experiments, while substantial, did not extend to the extreme scales of the largest industrial models, leaving the predictive accuracy at even higher orders of magnitude of compute as a subject for future work. Finally, our investigation into code-NL mixtures was focused on two distant points (70/30 and 30/70) to maximize observable differences. A more fine-grained study with additional ratios is needed to map the entire mixture spectrum and identify a more precise optimal blend.

\section*{Ethical Considerations}
The data used is from ~\citet{huang-etal-2025-opencoder,xuyang2025compressionreallylinearcode,Li2025Farseer} and others are solely from publicly accessible project resources on reputable websites, ensuring that no sensitive information is included. 
We have taken care to acknowledge the original authors by properly citing their work.

\section*{Acknowledge}
We gratefully acknowledge the support of the National Natural Science Foundation of China (NSFC) via grant 62236004 and 62476073.

\bibliography{ACL2026_conference}

\appendix

\section{Model Configurations}
\label{app:model_configs}

The architectural hyperparameters for a representative set of the 117 models are enumerated in Table~\ref{tab:detailed_model_configs}. In accordance with the methodology outlined in Section 2.3, each model's architecture was algorithmically derived from its target non-embedding parameter count ($N$). This deterministic mapping maintains structural consistency and near-optimal aspect ratios across scales, thereby isolating the effects of scale from architectural variance.

For the scaling-law experiments, training sets at different $D$ are obtained as nested subsets of the same pool. We shuffle the entire pool once with a fixed random seed and take progressively larger contiguous prefixes to form each budget. This design keeps the distribution stable across budgets.

To evaluate trained models, we measure loss on a high-quality validation set drawn from a closed-source internal codebase~\cite{xuyang2025compressionreallylinearcode}. It is written by experienced engineers for production use, which aims to ensure realism and quality. Its closed-source nature reduces the risk of contamination from the public training pool. We further perform exact and near-duplicate checks against the training pool to guard against accidental overlap. Although no single source can perfectly represent the entire distribution of code, this validation set provides a consistent benchmark with checks to minimize contamination across all models. The final validation set contains 6.3M tokens.

For data-mixing experiments, the NL train data we used is from ~\citet{Li2025Farseer}.

\section{Dataset and Validation Details}
\label{app:data_details}

We use the pretraining corpus released by OpenCoder~\citep{huang-etal-2025-opencoder} as the training set, which covers public GitHub content up to November 2023. The dataset has already undergone strict deduplication and rule-based filtering, following best practices established for large-scale open code corpora such as The Stack~\citep{lozhkov2024starcoder2}. The strong performance of OpenCoder indicates the coverage and quality of this corpus. We remove synthetic data to better match the natural distribution of real-world code.

After tokenization with our in-house tokenizer, the raw corpus contains 1216.93B tokens. Since knowledge density does not scale proportionally with raw volume across programming languages, we downsample languages with high redundancy and extreme volume to improve balance and pretraining efficiency. The resulting train set contains 895.51B tokens. Table~\ref{tab:train_set} reports the detailed language composition and summary statistics.

\begin{table*}[]
\centering
\begin{tabular}{lrrrr}
\toprule
\textbf{Language} & \textbf{Original (B)} & \textbf{Sample} & \textbf{Sampled (B)} & \textbf{Weight(\%)} \\
\midrule
C & 52.33 & 1.00 & 52.33 & 5.84 \\
C++ & 67.43 & 1.00 & 67.43 & 7.53 \\
C\# & 66.53 & 1.00 & 66.53 & 7.43 \\
Go & 12.76 & 1.00 & 12.76 & 1.43 \\
HTML & 260.27 & 0.05 & 13.01 & 1.45 \\
Java & 148.29 & 0.50 & 74.14 & 8.28 \\
JavaScript & 75.84 & 1.00 & 75.84 & 8.47 \\
Others & 307.22 & 1.00 & 307.22 & 34.31 \\
PHP & 75.67 & 1.00 & 75.67 & 8.45 \\
Python & 74.53 & 1.00 & 74.53 & 8.32 \\
Jupyter & 15.04 & 1.00 & 15.04 & 1.68 \\
Stack v2 & 55.46 & 1.00 & 55.46 & 6.19 \\
LeetCode & 5.56 & 1.00 & 5.56 & 0.62 \\
\midrule
\textbf{Total} & \textbf{1,216.93} & \textbf{--} & \textbf{895.51} & \textbf{100.00} \\
\bottomrule
\end{tabular}
\caption{Composition and sampling configuration of the training set by language.}
\label{tab:train_set}
\end{table*}

\section{Computational Setup}
\label{app:comp_setup}

All experiments are conducted on a cluster of NVIDIA H100-80GB GPUs. A meticulous resource allocation strategy is employed to maximize computational efficiency while adhering to the optimal training hyperparameters. For each run, the optimal global batch size (GBZ) is calculated by $N$ and $D$ through a compute-aware heuristic. The maximum per-GPU micro batch size (MBZ), is then determined based on the model's size and training sequence length. This value can be slightly lower in multi-GPU setups. While using more GPUs can accelerate experiments, it often reduces Model FLOPs Utilization (MFU). We carefully manage this trade-off.

The ideal number of GPUs, approximated by GBZ / MBZ, often requires adjustment due to hardware constraints such as integer divisibility. In making the adjustment, the priority is to maintain the GBZ as close to the optimal number as possible, given its critical impact on scaling law dynamics. Across our experimental runs, this strategy resulted in configurations using 8 to 128 GPUs, with a total compute equivalent to approximately 13,600 A100 GPU-days.

\section{Irreducible Loss Limit Derivation}
\label{app:irreducible_loss}

A key theoretical divergence between the Chinchilla and Farseer scaling laws lies in their prediction of the model's performance limit. The Chinchilla formulation posits a non-zero irreducible loss, $E$, while the Farseer law's structure allows for a limit that can be zero. This difference is a direct consequence of their mathematical structures. For our fitted Chinchilla law (Equation 3), the limit is a substantial non-zero constant:
\begin{equation}
\lim_{N,D \to \infty} \left( 0.2193 + \frac{534.374}{N^{0.4853}} + \frac{76.0743}{D^{0.2983}} \right) = 0.2193
\end{equation}

For our specific fitted Farseer law (Equation 4), while one term decays to zero, the second term converges to a very small, non-zero constant, making the final limit non-zero:
\begin{align}
\lim_{N,D \to \infty} L(N,D) &= \lim_{N \to \infty} e^{-0.0047 N^{0.239} - 0.8188} \nonumber \\
&\quad + \lim_{N,D \to \infty} e^{62.8936 N^{-0.0614} - 14.0414} \nonumber \\
&\quad \times D^{-e^{-0.0209 N^{0.1943} - 0.1826}} \nonumber \\
&= 0 + e^{-14.0414} \cdot \lim_{D \to \infty} D^{0} \nonumber \\
&= e^{-14.0414} \approx 8.00 \times 10^{-7}
\end{align}

While our specific fit results in a near-zero irreducible loss rather than a true zero, it still stands in stark contrast to the much larger value predicted by the Chinchilla model. This motivates a deeper look into the theoretical argument for why such a limit should be possible, as it explores why, for models with sufficient context, the empirical entropy of a finite dataset can be expected to approach zero.

From an information-theoretic perspective, this discussion is centered on the entropy rate $h$ of the data, which can be expressed as the limit of the conditional entropy:
\begin{equation}
h = \lim_{n \to \infty} H(X_n | X_1, \ldots, X_{n-1})
\end{equation}

The central question is whether $h$ must be a positive constant. While this may be true for an idealized, infinite data generating process, it is plausible to argue that the empirical entropy rate of any finite training corpus trends towards zero as the modeled context length $n$ becomes sufficiently large. To make this abstract concept more concrete, let us examine the practical realities of modern transformers.

Consider a model with a typical context window of $n=4096$ tokens. Its objective is to predict the $4096$-th token given the prefix of $4095$ tokens. The space of all possible prefixes is vast, on the order of $V^{4095}$ where $V$ is the vocabulary size. Given the finite size of any real-world training corpus (e.g., trillions of tokens), this number is minuscule compared to the space of possible prefixes. Consequently, it is highly improbable that a specific, long prefix of 4095 tokens appears more than once in the entire dataset.

For the vast majority of training instances, the model is therefore presented with a prefix that is empirically unique. This uniqueness implies that the token that follows it is also unique from the dataset's point of view, making it deterministic in this context. In such cases, the empirical conditional probability for the next token is effectively 1 for a single outcome. The conditional entropy for such a unique prefix is therefore zero:
\begin{align}
H(&X_{4096} | X_1, \ldots, X_{4095}) \nonumber \\
&= -\sum_{x} P(x|\text{prefix}) \log_2 P(x|\text{prefix}) \nonumber \\
&\approx -(1 \cdot \log_2 1) = 0
\end{align}

A unique prefix leading to a deterministic next token is not a rare edge case but the dominant scenario when dealing with large context windows on finite data. This prevalence of empirically deterministic sequences suggests that as a model's context capacity increases, the average empirical entropy it is tasked with modeling should decrease. Therefore, a scaling law whose functional form permits a limit of or near zero, like Farseer, may provide a more accurate theoretical foundation for transformers with large context windows. Such models have the capacity to leverage long-range, near-deterministic patterns that are inherent to any finite data collection.

\onecolumn

\begin{longtable}{ccccccccccc}
\toprule
\textbf{$N$(B)} & \textbf{$D$(B)} & \textbf{$h$} & \textbf{$ffnh$} & \textbf{$numl$} & \textbf{$numh$} & \textbf{$lr$} & \textbf{$gbz$} & \textbf{$iter$} & \textbf{$gpus$} & \textbf{$mbz\_max$} \\
\midrule
\endfirsthead

\toprule
\textbf{$N$(B)} & \textbf{$D$(B)} & \textbf{$h$} & \textbf{$ffnh$} & \textbf{$numl$} & \textbf{$numh$} & \textbf{$lr$} & \textbf{$gbz$} & \textbf{$iter$} & \textbf{$gpus$} & \textbf{$mbz\_max$} \\
\midrule
\endhead

\midrule
\endfoot

\bottomrule
\caption{Complete training configurations for all models.} \label{tab:detailed_model_configs}
\endlastfoot
0.201 & 2.00 & 1024 & 2728 & 16 & 16 & 0.001565951 & 56 & 17438 & 8 & 16 \\
0.284 & 2.00 & 1152 & 3032 & 18 & 18 & 0.001224257 & 56 & 17438 & 8 & 16 \\
0.398 & 2.00 & 1280 & 3472 & 20 & 20 & 0.000963343 & 56 & 17438 & 8 & 16 \\
0.568 & 2.00 & 1472 & 3888 & 22 & 23 & 0.000746831 & 56 & 17438 & 8 & 16 \\
0.798 & 2.00 & 1600 & 4264 & 26 & 25 & 0.000586174 & 56 & 17438 & 8 & 16 \\
1.130 & 2.00 & 1792 & 4832 & 29 & 28 & 0.000458791 & 56 & 17438 & 8 & 8 \\
1.610 & 2.00 & 2048 & 5448 & 32 & 32 & 0.000355834 & 56 & 17438 & 8 & 8 \\
2.270 & 2.00 & 2304 & 6064 & 36 & 36 & 0.000277997 & 56 & 17438 & 8 & 8 \\
3.180 & 2.00 & 2560 & 6952 & 40 & 40 & 0.00021863 & 56 & 17438 & 8 & 8 \\
0.201 & 2.83 & 1024 & 2728 & 16 & 16 & 0.001742048 & 72 & 19181 & 8 & 16 \\
0.284 & 2.83 & 1152 & 3032 & 18 & 18 & 0.00136193 & 72 & 19181 & 8 & 16 \\
0.398 & 2.83 & 1280 & 3472 & 20 & 20 & 0.001071675 & 72 & 19181 & 8 & 16 \\
0.568 & 2.83 & 1472 & 3888 & 22 & 23 & 0.000830816 & 72 & 19181 & 8 & 16 \\
0.798 & 2.83 & 1600 & 4264 & 26 & 25 & 0.000652091 & 72 & 19181 & 8 & 16 \\
1.130 & 2.83 & 1792 & 4832 & 29 & 28 & 0.000510384 & 72 & 19181 & 8 & 8 \\
1.610 & 2.83 & 2048 & 5448 & 32 & 32 & 0.000395849 & 72 & 19181 & 8 & 8 \\
2.270 & 2.83 & 2304 & 6064 & 36 & 36 & 0.000309259 & 72 & 19181 & 8 & 8 \\
3.180 & 2.83 & 2560 & 6952 & 40 & 40 & 0.000243215 & 72 & 19181 & 8 & 8 \\
0.201 & 4.00 & 1024 & 2728 & 16 & 16 & 0.001937948 & 88 & 22194 & 8 & 16 \\
0.284 & 4.00 & 1152 & 3032 & 18 & 18 & 0.001515084 & 88 & 22194 & 8 & 16 \\
0.398 & 4.00 & 1280 & 3472 & 20 & 20 & 0.001192189 & 88 & 22194 & 8 & 16 \\
0.568 & 4.00 & 1472 & 3888 & 22 & 23 & 0.000924244 & 88 & 22194 & 8 & 16 \\
0.798 & 4.00 & 1600 & 4264 & 26 & 25 & 0.000725421 & 88 & 22194 & 8 & 16 \\
1.130 & 4.00 & 1792 & 4832 & 29 & 28 & 0.000567778 & 88 & 22194 & 8 & 8 \\
1.610 & 4.00 & 2048 & 5448 & 32 & 32 & 0.000440364 & 88 & 22194 & 8 & 8 \\
2.270 & 4.00 & 2304 & 6064 & 36 & 36 & 0.000344036 & 88 & 22194 & 8 & 8 \\
3.180 & 4.00 & 2560 & 6952 & 40 & 40 & 0.000270566 & 88 & 22194 & 8 & 8 \\
0.201 & 5.66 & 1024 & 2728 & 16 & 16 & 0.002155878 & 104 & 26558 & 8 & 16 \\
0.284 & 5.66 & 1152 & 3032 & 18 & 18 & 0.001685461 & 104 & 26558 & 8 & 16 \\
0.398 & 5.66 & 1280 & 3472 & 20 & 20 & 0.001326255 & 104 & 26558 & 8 & 16 \\
0.568 & 5.66 & 1472 & 3888 & 22 & 23 & 0.001028179 & 104 & 26558 & 8 & 16 \\
0.798 & 5.66 & 1600 & 4264 & 26 & 25 & 0.000806998 & 104 & 26558 & 8 & 16 \\
1.130 & 5.66 & 1792 & 4832 & 29 & 28 & 0.000631627 & 104 & 26558 & 8 & 8 \\
1.610 & 5.66 & 2048 & 5448 & 32 & 32 & 0.000489885 & 104 & 26558 & 8 & 8 \\
2.270 & 5.66 & 2304 & 6064 & 36 & 36 & 0.000382724 & 104 & 26558 & 8 & 8 \\
3.180 & 5.66 & 2560 & 6952 & 40 & 40 & 0.000300992 & 104 & 26558 & 8 & 8 \\
0.201 & 8.00 & 1024 & 2728 & 16 & 16 & 0.002398315 & 128 & 30517 & 8 & 16 \\
0.284 & 8.00 & 1152 & 3032 & 18 & 18 & 0.001874997 & 128 & 30517 & 8 & 16 \\
0.398 & 8.00 & 1280 & 3472 & 20 & 20 & 0.001475398 & 128 & 30517 & 8 & 16 \\
0.568 & 8.00 & 1472 & 3888 & 22 & 23 & 0.001143801 & 128 & 30517 & 8 & 16 \\
0.798 & 8.00 & 1600 & 4264 & 26 & 25 & 0.000897748 & 128 & 30517 & 8 & 16 \\
1.130 & 8.00 & 1792 & 4832 & 29 & 28 & 0.000702656 & 128 & 30517 & 8 & 8 \\
1.610 & 8.00 & 2048 & 5448 & 32 & 32 & 0.000544974 & 128 & 30517 & 8 & 8 \\
2.270 & 8.00 & 2304 & 6064 & 36 & 36 & 0.000425763 & 128 & 30517 & 8 & 8 \\
3.180 & 8.00 & 2560 & 6952 & 40 & 40 & 0.00033484 & 128 & 30517 & 8 & 8 \\
0.201 & 11.30 & 1024 & 2728 & 16 & 16 & 0.002668015 & 152 & 36343 & 8 & 16 \\
0.284 & 11.30 & 1152 & 3032 & 18 & 18 & 0.002085848 & 152 & 36343 & 8 & 16 \\
0.398 & 11.30 & 1280 & 3472 & 20 & 20 & 0.001641312 & 152 & 36343 & 8 & 16 \\
0.568 & 11.30 & 1472 & 3888 & 22 & 23 & 0.001272426 & 152 & 36343 & 8 & 16 \\
0.798 & 11.30 & 1600 & 4264 & 26 & 25 & 0.000998703 & 152 & 36343 & 8 & 16 \\
1.130 & 11.30 & 1792 & 4832 & 29 & 28 & 0.000781672 & 152 & 36343 & 8 & 8 \\
1.610 & 11.30 & 2048 & 5448 & 32 & 32 & 0.000606259 & 152 & 36343 & 8 & 8 \\
2.270 & 11.30 & 2304 & 6064 & 36 & 36 & 0.000473641 & 152 & 36343 & 8 & 8 \\
3.180 & 11.30 & 2560 & 6952 & 40 & 40 & 0.000372494 & 160 & 34526 & 16 & 8 \\
0.201 & 16.00 & 1024 & 2728 & 16 & 16 & 0.002968043 & 192 & 40690 & 8 & 16 \\
0.284 & 16.00 & 1152 & 3032 & 18 & 18 & 0.002320409 & 192 & 40690 & 8 & 16 \\
0.398 & 16.00 & 1280 & 3472 & 20 & 20 & 0.001825884 & 192 & 40690 & 8 & 16 \\
0.568 & 16.00 & 1472 & 3888 & 22 & 23 & 0.001415515 & 192 & 40690 & 8 & 16 \\
0.798 & 16.00 & 1600 & 4264 & 26 & 25 & 0.001111011 & 192 & 40690 & 8 & 16 \\
1.130 & 16.00 & 1792 & 4832 & 29 & 28 & 0.000869574 & 192 & 40690 & 8 & 8 \\
1.610 & 16.00 & 2048 & 5448 & 32 & 32 & 0.000674435 & 192 & 40690 & 8 & 8 \\
2.270 & 16.00 & 2304 & 6064 & 36 & 36 & 0.000526904 & 192 & 40690 & 16 & 8 \\
3.180 & 16.00 & 2560 & 6952 & 40 & 40 & 0.000414382 & 192 & 40690 & 16 & 8 \\
0.201 & 22.60 & 1024 & 2728 & 16 & 16 & 0.003301811 & 232 & 47623 & 8 & 16 \\
0.284 & 22.60 & 1152 & 3032 & 18 & 18 & 0.002581348 & 232 & 47623 & 8 & 16 \\
0.398 & 22.60 & 1280 & 3472 & 20 & 20 & 0.002031211 & 232 & 47623 & 8 & 16 \\
0.568 & 22.60 & 1472 & 3888 & 22 & 23 & 0.001574696 & 232 & 47623 & 8 & 16 \\
0.798 & 22.60 & 1600 & 4264 & 26 & 25 & 0.001235948 & 232 & 47623 & 8 & 16 \\
1.130 & 22.60 & 1792 & 4832 & 29 & 28 & 0.000967361 & 232 & 47623 & 8 & 8 \\
1.610 & 22.60 & 2048 & 5448 & 32 & 32 & 0.000750278 & 224 & 49323 & 16 & 8 \\
2.270 & 22.60 & 2304 & 6064 & 36 & 36 & 0.000586157 & 224 & 49323 & 16 & 8 \\
3.180 & 22.60 & 2560 & 6952 & 40 & 40 & 0.000460981 & 224 & 49323 & 32 & 8 \\
0.201 & 32.00 & 1024 & 2728 & 16 & 16 & 0.003673112 & 280 & 55803 & 8 & 16 \\
0.284 & 32.00 & 1152 & 3032 & 18 & 18 & 0.002871631 & 280 & 55803 & 8 & 16 \\
0.398 & 32.00 & 1280 & 3472 & 20 & 20 & 0.002259629 & 280 & 55803 & 8 & 16 \\
0.568 & 32.00 & 1472 & 3888 & 22 & 23 & 0.001751776 & 280 & 55803 & 8 & 16 \\
0.798 & 32.00 & 1600 & 4264 & 26 & 25 & 0.001374935 & 280 & 55803 & 8 & 16 \\
1.130 & 32.00 & 1792 & 4832 & 29 & 28 & 0.001076145 & 288 & 54253 & 16 & 8 \\
1.610 & 32.00 & 2048 & 5448 & 32 & 32 & 0.000834649 & 288 & 54253 & 16 & 8 \\
2.270 & 32.00 & 2304 & 6064 & 36 & 36 & 0.000652072 & 288 & 54253 & 32 & 8 \\
3.180 & 32.00 & 2560 & 6952 & 40 & 40 & 0.00051282 & 288 & 54253 & 32 & 8 \\
0.201 & 45.30 & 1024 & 2728 & 16 & 16 & 0.004086168 & 344 & 64235 & 8 & 16 \\
0.284 & 45.30 & 1152 & 3032 & 18 & 18 & 0.003194557 & 344 & 64235 & 8 & 16 \\
0.398 & 45.30 & 1280 & 3472 & 20 & 20 & 0.002513733 & 344 & 64235 & 8 & 16 \\
0.568 & 45.30 & 1472 & 3888 & 22 & 23 & 0.00194877 & 344 & 64235 & 8 & 16 \\
0.798 & 45.30 & 1600 & 4264 & 26 & 25 & 0.001529552 & 336 & 65765 & 16 & 16 \\
1.130 & 45.30 & 1792 & 4832 & 29 & 28 & 0.001197162 & 336 & 65765 & 16 & 8 \\
1.610 & 45.30 & 2048 & 5448 & 32 & 32 & 0.000928509 & 352 & 62775 & 32 & 8 \\
2.270 & 45.30 & 2304 & 6064 & 36 & 36 & 0.0007254 & 352 & 62775 & 32 & 8 \\
3.180 & 45.30 & 2560 & 6952 & 40 & 40 & 0.000570489 & 352 & 62775 & 32 & 8 \\
0.201 & 64.00 & 1024 & 2728 & 16 & 16 & 0.004545673 & 416 & 75120 & 8 & 16 \\
0.284 & 64.00 & 1152 & 3032 & 18 & 18 & 0.003553797 & 416 & 75120 & 8 & 16 \\
0.398 & 64.00 & 1280 & 3472 & 20 & 20 & 0.002796412 & 416 & 75120 & 8 & 16 \\
0.568 & 64.00 & 1472 & 3888 & 22 & 23 & 0.002167917 & 416 & 75120 & 16 & 16 \\
0.798 & 64.00 & 1600 & 4264 & 26 & 25 & 0.001701556 & 416 & 75120 & 16 & 16 \\
1.130 & 64.00 & 1792 & 4832 & 29 & 28 & 0.001331787 & 416 & 75120 & 32 & 8 \\
1.610 & 64.00 & 2048 & 5448 & 32 & 32 & 0.001032923 & 416 & 75120 & 32 & 8 \\
2.270 & 64.00 & 2304 & 6064 & 36 & 36 & 0.000806974 & 448 & 69754 & 64 & 8 \\
3.180 & 64.00 & 2560 & 6952 & 40 & 40 & 0.000634642 & 448 & 69754 & 64 & 8 \\
0.201 & 90.50 & 1024 & 2728 & 16 & 16 & 0.005056852 & 512 & 86316 & 8 & 16 \\
0.284 & 90.50 & 1152 & 3032 & 18 & 18 & 0.003953435 & 512 & 86316 & 16 & 16 \\
0.398 & 90.50 & 1280 & 3472 & 20 & 20 & 0.003110879 & 512 & 86316 & 16 & 16 \\
0.568 & 90.50 & 1472 & 3888 & 22 & 23 & 0.002411707 & 512 & 86316 & 16 & 16 \\
0.798 & 90.50 & 1600 & 4264 & 26 & 25 & 0.001892903 & 512 & 86316 & 32 & 16 \\
1.130 & 90.50 & 1792 & 4832 & 29 & 28 & 0.001481551 & 512 & 86316 & 32 & 8 \\
1.610 & 90.50 & 2048 & 5448 & 32 & 32 & 0.001149079 & 512 & 86316 & 64 & 8 \\
2.270 & 90.50 & 2304 & 6064 & 36 & 36 & 0.000897722 & 512 & 86316 & 64 & 8 \\
3.180 & 90.50 & 2560 & 6952 & 40 & 40 & 0.00070601 & 512 & 86316 & 64 & 4 \\
0.201 & 128.00 & 1024 & 2728 & 16 & 16 & 0.005625514 & 624 & 100160 & 16 & 16 \\
0.284 & 128.00 & 1152 & 3032 & 18 & 18 & 0.004398014 & 624 & 100160 & 16 & 16 \\
0.398 & 128.00 & 1280 & 3472 & 20 & 20 & 0.003460709 & 624 & 100160 & 16 & 16 \\
0.568 & 128.00 & 1472 & 3888 & 22 & 23 & 0.002682913 & 608 & 102796 & 32 & 16 \\
0.798 & 128.00 & 1600 & 4264 & 26 & 25 & 0.002105767 & 608 & 102796 & 32 & 16 \\
1.130 & 128.00 & 1792 & 4832 & 29 & 28 & 0.001648158 & 640 & 97656 & 64 & 8 \\
1.610 & 128.00 & 2048 & 5448 & 32 & 32 & 0.001278297 & 640 & 97656 & 64 & 8 \\
2.270 & 128.00 & 2304 & 6064 & 36 & 36 & 0.000998674 & 640 & 97656 & 64 & 8 \\
3.180 & 128.00 & 2560 & 6952 & 40 & 40 & 0.000785404 & 640 & 97656 & 64 & 8 \\

\end{longtable}

\twocolumn

\end{document}